\documentclass[letterpaper, 10 pt, conference]{ieeeconf}  

\IEEEoverridecommandlockouts                              

\usepackage{amsmath,amssymb,amsfonts}
\usepackage{algorithm}
\usepackage{algpseudocode}
\usepackage{graphicx}
\usepackage{textcomp}
\usepackage{xcolor}
\usepackage{comment}
\usepackage{textcomp}
\usepackage{amsfonts}
\usepackage{booktabs}
\usepackage{array}
\usepackage{multirow}
\usepackage{afterpage}
\usepackage{bm}
\usepackage[hidelinks]{hyperref}

\title{\LARGE \bf
Enhanced Human-Robot Collaboration using Constrained Probabilistic Human-Motion Prediction}

\author{Aadi Kothari$^{1}$ \thanks{$^{1}$The authors are with the Department of Mechanical Engineering at the Massachusetts Institute of Technology. E-mail: \tt\small aadi@mit.edu, tohme@mit.edu, kevxt@mit.edu, youcef@mit.edu}, Tony Tohme$^{1}$, Xiaotong Zhang$^{1}$, and Kamal Youcef-Toumi$^{1}$
}

\begin{document}

\maketitle
\thispagestyle{empty}
\pagestyle{empty}

\begin{abstract}

Human motion prediction is an essential step for efficient and safe human-robot collaboration. Current methods either purely rely on representing the human joints in some form of neural network-based architecture or use regression models offline to fit hyper-parameters in the hope of capturing a model encompassing human motion. While these methods provide good initial results, they are missing out on leveraging well-studied human body kinematic models as well as body and scene constraints which can help boost the efficacy of these prediction frameworks while also explicitly avoiding implausible human joint configurations. We propose a novel human motion prediction framework that incorporates human joint constraints and scene constraints in a Gaussian Process Regression (GPR) model to predict human motion over a set time horizon. This formulation is combined with an online context-aware constraints model to leverage task-dependent motions. It is tested on a human arm kinematic model and implemented on a human-robot collaborative setup with a UR5 robot arm to demonstrate the real-time capability of our approach. Simulations were also performed on datasets like HA4M and ANDY. The simulation and experimental results demonstrate considerable improvements in a Gaussian Process framework when these constraints are explicitly considered.

\end{abstract}

\section{INTRODUCTION}

Motion prediction is an essential step for efficient and safe human-robot collaboration. Often times there are tasks that necessitate human expertise and robotic precision, requiring varying degrees of collaboration between the two. Humans often tend to anticipate each other's motion to avoid collisions while efficiently achieving short-term and long-term objectives in a collaborative setting. Thus, it is crucial to predict human motion for robots to navigate safely around humans while also making sure that the planned motion is efficient in space and time. 

Among the numerous approaches to human motion prediction, some still result in improbable or impractical predictions, leading to lower prediction accuracy.
Methods ranging from model-based methods that exploit the inherent dynamics of physical systems \cite{simulating_human_motion}, to approaches like Inverse Optimal Control (IOC) \cite{ioc1, ioc2}, which seek to emulate cost functions that rationalize observed movements, have been attempted to make accurate prediction. In more recent developments, data-driven strategies have emerged that try to decipher not only the underlying dynamics but also the intricate causal relationships with the environment, contingent on the task at hand. Yet, for the real-world application of these techniques, it is necessary to consider sensor noise, an inherent pain point when tracking human joints, especially for markerless methods. In doing so, the propagation of this uncertainty into subsequent prediction stages becomes important. Regrettably, the current techniques that are capable of quantifying this uncertainty \cite{prob_diffusion, task_constrained_mp} often omit the critical aspect of using the kinematics of the physical system at hand and also fail to account for the context of the workspace within which it operates. The neglect of these factors leads to an unadjusted probability distribution function (pdf) for predictions. This underlines the pivotal role of accounting for physical systems, their intrinsic constraints, and workspace limitations when handling prediction uncertainty. Disregarding them can result in models generating improbable or impractical predictions, despite their mathematical feasibility, thus underscoring the imperative to integrate these elements for a more practical motion prediction model.

Uncertainty propagation from pose estimation to prediction is essential for accurate motion prediction of a physical system. This uncertainty can also be propagated in the downstream layers of a robotics stack \cite{risk_bounded}. Human joints are well-studied and it is important to make use of the kinematic constraints of the body as well as the physical constraints of the operating space for making accurate predictions.  For accurate human motion prediction with sufficient representability, not only do we need to track action based on past motion of human joints \cite{iplanner}, which inherently has uncertainty associated with it, we need to propagate this uncertainty through the prediction layer while making sure that the probability distribution function of the prediction satisfies kinematic and workspace constraints.

\begin{figure*}[htbp]
\vspace{0.1in}
\centerline{\includegraphics[width=16.0cm, height=5.05cm]{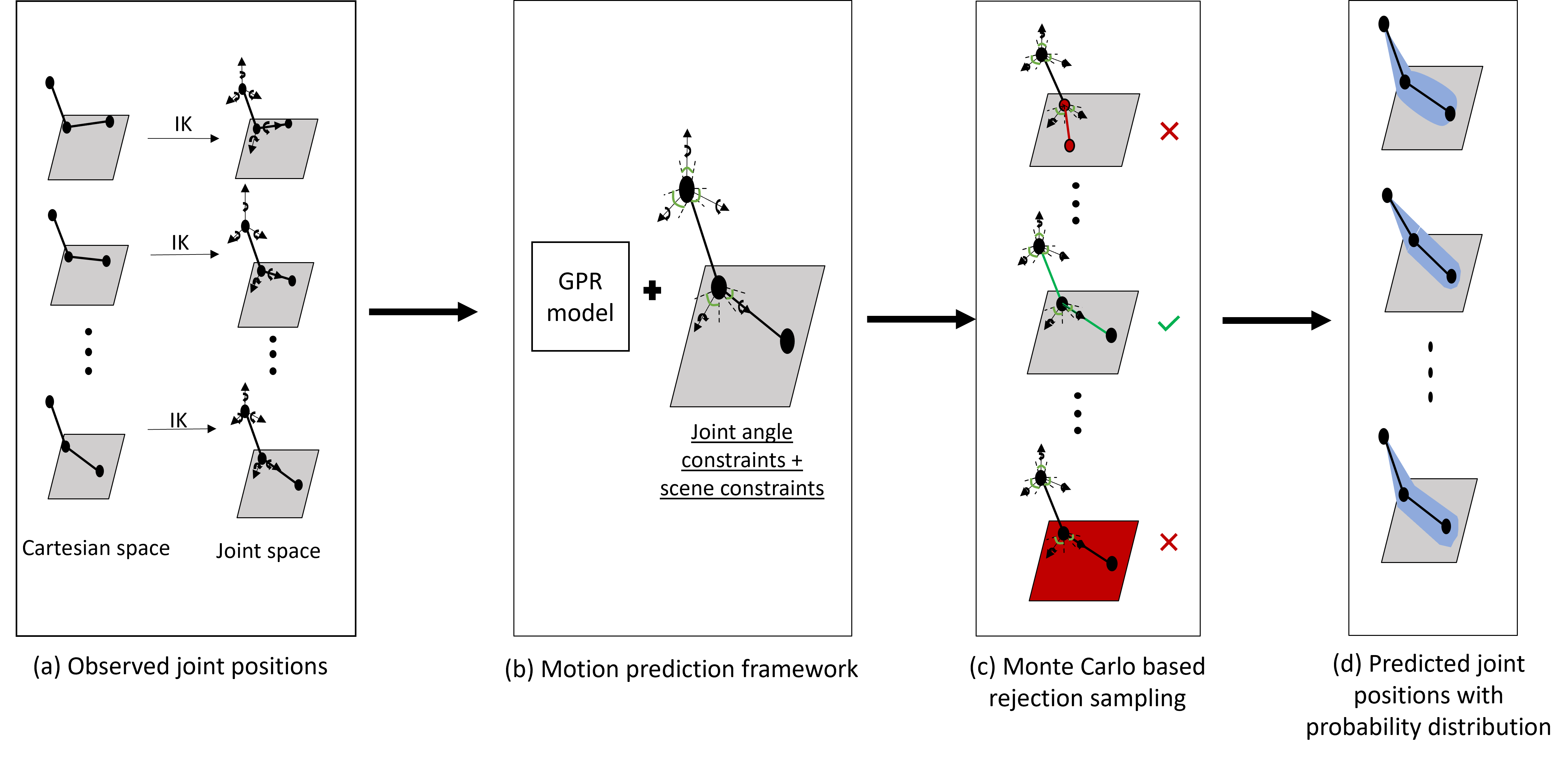}}
\caption{Overview of the proposed constrained probability distribution prediction for human motion. Inverse kinematics is used to get to joint space for each observed time step, passed through a GPR model with constraints which are imposed using rejection sampling. Red joints represent a violation of joint angle or velocity constraints while red plane denotes collision or intersection of the prediction with an object in the scene.}
\label{fig_flow_chart}
\end{figure*}

The contributions of this paper are as follows. A framework for human motion prediction is proposed that takes measurement uncertainty into account to predict future motion with corresponding uncertainty while respecting the kinematic and physical constraints of the human body. Further, this uncertainty is propagated into the task space and the predicted probability distribution function is modified to respect task space constraints. An overview of the framework is shown in Fig. \ref{fig_flow_chart}. This formulation not only helps in getting rid of implausible predictions that violate human joint configurations but also incorporates context into the uncertainty framework. Our framework is evaluated on two different human motion datasets, demonstrating the benefit of considering such constraints on the output pdf. Lastly, we implement it on a human-robot collaborative setup, where we use human motion predictions to make informed decisions by a UR5 robot arm.

\section{Related Work}

\subsection{Uncertainty quantification in motion prediction}

Current methods do not explicitly take human body joint characteristics, and contextual physical constraints into account, especially when these constraints operate in different representation spaces. Gaussian Process Dynamical Models (GPDM) \cite{gpdm} have been a traditional method for modelling human dynamics in order to predict human motion while also quantifying uncertainty. State-of-the-art deep learning architectures have also been recently modified for uncertainty quantification \cite{task_constrained_mp}, or work like \cite{reliable_nn} attempts to purely quantify regression-based uncertainty. Unfortunately, none of them account for constraints in their framework to avoid predicting implausible configurations.

\subsection{Constrained Gaussian Process Regression}

For human motion prediction, we can enforce several different kinds of constraints like position, velocity, etc. constraints when operating in joint space and collision constraints with physical objects when operating in the task space. As for a GPR problem formulation, the constraints can be bound constraints, monotonicity and convexity constraints, as well as differential equation constraints \cite{constr_gpr}. 
In the realm of GPR, warping functions \cite{warped_gp}, as well as non-gaussian likelihood functions are among approaches that enforce bound constraints. When considering non-gaussian likelihood functions to enforce constraints, the posterior is often analytically intractable, and truncated Gaussian distributions have been considered instead to enforce discrete constraints \cite{constr_gpr}. We leverage truncated normal distribution along with rejection sampling for our framework. 

\subsection{Model-based human motion prediction}
It is important to leverage joint angle position, velocity, etc. constraints to get rid of implausible human motion predictions. Kinematics and dynamics modelling has traditionally been an important component for understanding how humans move. The human body has been well studied and the nature of kinematic chains and physics-based methods have been leveraged in various models \cite{simulating_human_motion, physics_infused_nn}. Specifically, for human motion prediction, there are different representations which can be used as a starting point for the prediction model, the simplest one being the x,y,z cartesian coordinates of each joint at a given time. In recent literature \cite{st_transformer}, human joints representation like 3D position, rotation matrix, angle-axis, or quaternions have been considered. For our formulation, it can be assumed that the motion is independent in the joint angle space, i.e. movement of one joint does not inherently affect movement of another, since it helps in dimensionality reduction while leaving room for imposing constraints. The joint angle space, as part of the kinematic model, is thus a feasible representation as a starting point of our prediction framework. 

\subsection{Context-based human motion prediction}
Human motion prediction is context driven and it is necessary to include scene context within the prediction framework. Context-based interactions have been attempted and addressed in \cite{forecasting_in_wild, context_based_prediction}. Though these methods, via incorporating different neural network-based architectures like Graph Neural Networks \cite{sts_gcn}, spatio-temporal transformers \cite{st_transformer}, etc., are expected to inherently account for the physical attributes of a scene and collision avoidance in prediction, they lack elements that explicitly enforce context-based constraints in the prediction and also the corresponding uncertainty that should be modified accordingly. We take care of physical scene-based constraints using a rejection sampling-based approach in our framework.

\section{Problem formulation}
The goal of our approach is to make accurate human motion predictions while considering associated measurement uncertainty along with respecting kinematic and contextual constraints. We formulate the human motion prediction problem as an output of a GPR model in which we infuse the human kinematic model as well as physical workspace constraints. Similar to \cite{st_transformer}, we consider a sequence of observed motion as $\textbf{S}(t) = \{\textbf{s}_1, . . . , \textbf{s}_{T_{O}} \}$ where a frame $\textbf{s}_t = \{\textbf{j}^{(1)}_t, . . . , \textbf{j}^{(N)}_t\}$ denotes a pose at time step t with joints $\textbf{j}^{(n)}_t \in R^q$, where $q$ is the number of pose parameters, example: x,y,z, axis-angle representation, etc, and $N$ is the number of joints tracked at every instant. These measurements are considered noisy and for an observed time horizon $T_{O}$. The goal of the prediction model is to generate an accurate joint probability distribution for the predicted \{$\hat{s}_{T_{O}+1}, ... \hat{s}_{T_{O}+T_{P}}\}$ where $T_P$ is the prediction time horizon. We use a sliding window prediction framework, i.e. to predict $\hat{s}_{T_O+2}$, we use the sequence $\{\textbf{s}_2, ....., \textbf{s}_{T_O}, \hat{\textbf{s}}_{T_O+1} \}$. 

\subsection{Inverse kinematics and joint angle space}\label{AA}

We use inverse kinematics (IK) to transform the measured cartesian xyz space information to joint space. Given a pair of joint poses $(\textbf{j}^{(n)}_{t}, \textbf{j}^{(n+1)}_{t})$ in a kinematic chain for a given time step t, by defining the kinematic transformations and using forward kinematics, we can get the relation between the two joint poses as  
\begin{equation}
    \textbf{j}^{(n+1)}_{t} = H_1...H_{p_n} \textbf{j}^{(n)}_{t} = H \textbf{j}^{(n)}_{t}
\end{equation}

where each $H_i$ denotes a Denavit-Hartenberg transformation matrix \cite{denavit_hartenberg} which is a function of every $\theta$ in $\bm{\theta}_t = \{\theta_{1,t}, \theta_{2,t}...\theta_{p_n,t}\} \in R^{p_n}$ for $p_n$ number of joint angles that define the transformations from $\textbf{j}^{(n)}_{t}$ to $\textbf{j}^{(n+1)}_{t}$.

\subsection{Constrained Gaussian Process Regression Models}


Gaussian process regression models can be used to propagate uncertainty from one time step to the next. The human body is well studied, and in the joint space, joint angles have linear constraints as well as higher-order derivative constraints that should be accounted for when making predictions on their motion \cite{human_arm_kinematics}. To incorporate these into our prediction framework and account for the propagation of measurement noise across the model, for a given estimated $\hat{\theta}$ $\in$ $\bm{\theta}_t$ we assume a truncated normal distribution prior to each joint angle defined as: 

\begin{equation} \label{eq:tn}
    \hat{\theta} \sim \mathcal{TN}(\mu_{\theta}, \sigma_{\theta}^2, \theta_{{lb}}, \theta_{{ub}})
\end{equation}

where $ \theta_{{lb}}$ and $ \theta_{{ub}}$ are lower and upper bounds on $\theta$ and $\mu_{\theta}, \sigma_{\theta}^2$ are mean and variances obtained from the output of a Sparse Pseudo-input Gaussian Process (SPGP) model \cite{spgp}. More accurate representation of $\theta$ can be estimated from methods like \cite{messy}. We define the SPGP as GP (Gaussian Process):

\begin{equation} \label{eq:gp}
    f_{\hat{\theta}} \sim GP(m, K)
\end{equation}

where $m$ and $K$ are learned mean and covariance functions of the chosen representation space and dataset or set of observed trajectories, and $\theta_{{lb}}$ and $\theta_{{ub}}$ can be defined as:

\begin{equation}
\begin{aligned}
    \theta_{{lb}}  = \max(\theta_{{lb}}, \theta_{t-1} - \dot{\theta}_{{ub}} \Delta t) \\
    \theta_{{ub}}  = \min(\theta_{{ub}}, \theta_{t-1} + \dot{\theta}_{{ub}} \Delta t) \\
\end{aligned}
\end{equation}

where $\dot{\theta}_{{ub}}$ is the upper bound on angular velocity and the latter part of the max and min functions comes from a simple linear interpolation using maximum permissible joint velocity over a time step $\Delta t$, and thus we get the bounds for our truncated normal distribution.

\subsection{Probability propagation and task space constraints}
\begin{figure}[htbp]

\centerline{\includegraphics[width=9cm, height=6.8cm]{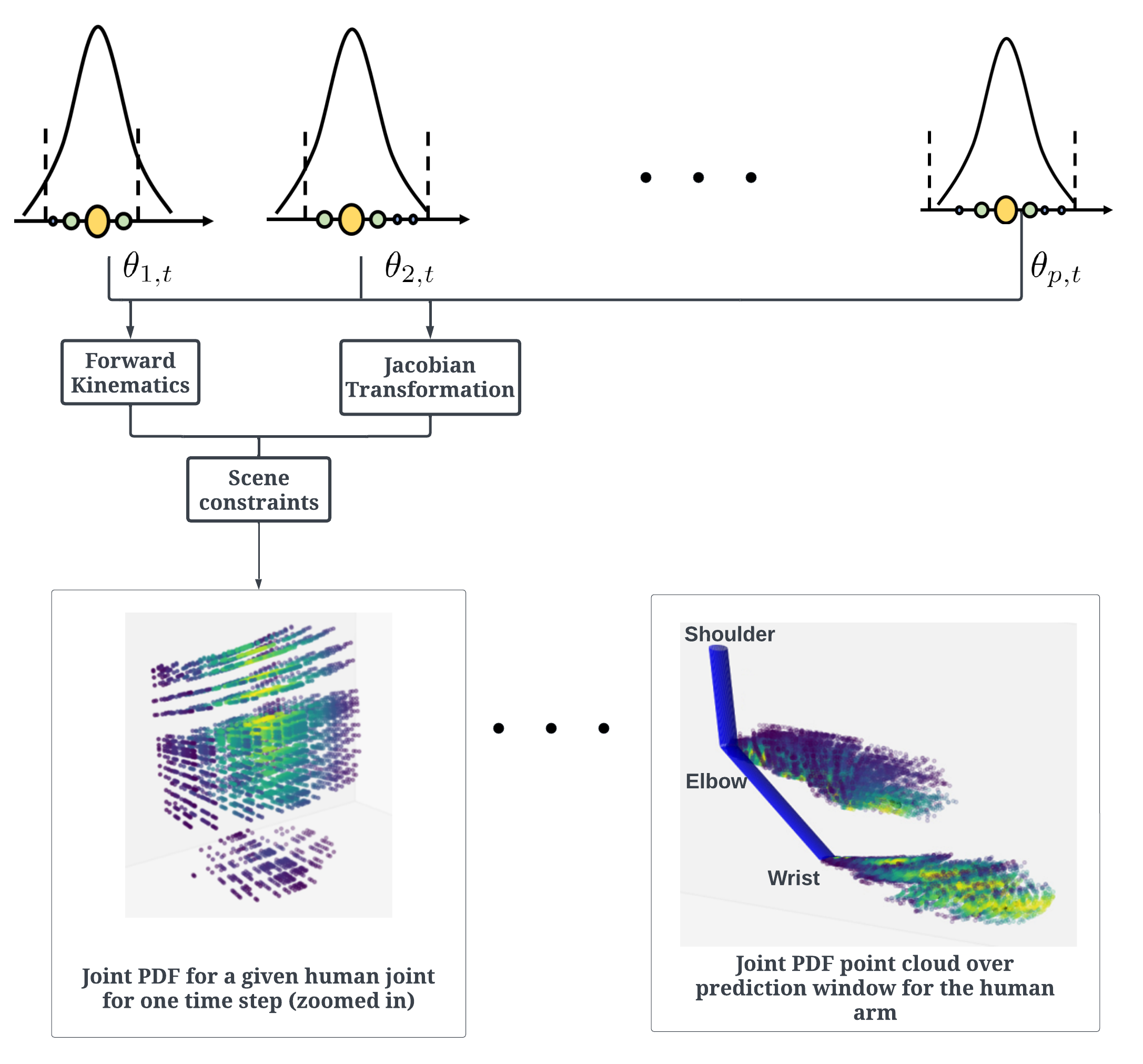}}
\caption{Constrained satisfaction-based Monte Carlo rejection sampling.}
\label{fig_mc}
\end{figure}

 We intend to propagate the probability distributions from joint space back to task space since we are operating in task space $R^3$. We can use the Jacobian transformation to transform individual predicted probability distribution functions in joint angles to joint position probability distributions. Firstly, we can define the Jacobian as:

 \begin{equation}
J = 
    \begin{bmatrix}
        \frac{\partial{X}}{\partial\theta_{1}} & \frac{\partial{X}}{\partial\theta_{2}} & \cdots & \frac{\partial{X}}{\partial\theta_{p_n}} \\[1ex]

        \frac{\partial{Y}}{\partial\theta_{1}} & \frac{\partial{Y}}{\partial\theta_{2}} & \cdots & \frac{\partial{Y}}{\partial\theta_{p_n}}
        \\[1ex]

        \frac{\partial{Z}}{\partial\theta_{1}} & \frac{\partial{Z}}{\partial\theta_{2}} & \cdots & \frac{\partial{Z}}{\partial\theta_{p_n}}
        \\
    \end{bmatrix}
\end{equation}

where $J \in R^{3 \times p_n}$. For a given time t, We define $f_{X_jY_jZ_j}$ as the joint probability distribution function in the workspace: 

\begin{equation} \label{eq:pdf}
    f_{X_jY_jZ_j} = \frac{f_{\theta_1}...f_{\theta_{p_n}}}{|J|}
\end{equation}

since our underlying assumption is that joint angle motions are independent with respect to each other. We note that for certain kinematic chain configurations, the Jacobian $J$ might not be a square matrix in which case, special methods exist to handle such issues like using pseudo determinants, singular value decomposition, etc. We can augment $J$ by introducing dummy variables and integrating over them. Since calculating this integral may not always be feasible, Monte Carlo (MC) based samples from Eq. (\ref{eq:tn}) can be used to further calculate these integrals using Reimann sums.
Since the workspace itself can have physical constraints due to the possible existence of objects, etc. in the scene, it is important to modify $f_{X_jY_jZ_j}$ to account for inequality constraints:

\begin{equation} \label{eq:xyz_constr}
\begin{aligned}
    x_{t, \min} &\leq x_t \leq x_{t, \max} \\
    y_{t, \min} &\leq y_t \leq y_{t, \max} \\
    z_{t, \min} &\leq z_t \leq z_{t, \max} \\
\end{aligned}
\end{equation}
 in the workspace coordinate frame where the minimum and maximum values for the constraints can be defined using static and dynamic obstacles in a workspace for a given prediction time $t$. The variables $x_t, y_t, z_t$ can be obtained from cartesian-like representation from $j_{t}^{(n+1)}$ using the MC samples and rejecting them when Eq. (\ref{eq:xyz_constr}) is not satisfied. A visual representation of this MC based rejection sampling approach leading to a joint pdf of the prediction is shown in Fig. \ref{fig_mc}. We normalize the resulting modified pdf using the sum of valid pdf values.
 
We note that our constrained human motion prediction framework, as highlighted in Algorithm \ref{alg:cap}, can be applied to any human motion prediction model that operates in joint space and can provide unconstrained uncertainty quantification of the predicted joint angle trajectories. Depending on the requirements of an application, and the level the sophistication desired on the modelling side, this model can be derived from Gaussian Processes, neural network-based models like spatio-temporal transformers \cite{st_transformer}, diffusion methods \cite{prob_diffusion}, etc. with an added output of uncertainty quantification.

\begin{algorithm}
\caption{Constrained human motion prediction using Gaussian Process Regression for a pair of joints for one prediction time horizon}\label{alg:cap}
\begin{algorithmic}[1]
\State \textbf{Offline}:
        \\
        \Comment Tune hyperparameters of SPGP based on existing observed trajectories or datasets.
        \State \textbf{Online:}
        \State Input: $S(t)$ \Comment{Observed motion sequence}
        \State $\theta_t = IK(j^{(n)}_{t}, j^{(n+1)}_{t})$ \Comment{Obtain joint angles using inverse kinematics}
        \For{t = $T_{O}$ to $T_{P}$ }
            \For{i = 1 to $p_n$}
                \State $u_{\hat{\theta}_{t+1}^{(i)}}, \sigma_{\hat{\theta}_{t+1}^{(i)}} \leftarrow \mathcal{GP}(\theta_t^{(i)}, \theta_{t-1}^{(i)}, \theta_{t-2}^{(i)}...)$ \Comment{GP one step sliding window prediction of mean, and variance}
                \State $\hat{\theta}_{t+1}^{(i)} \sim \mathcal{TN}(u_{\hat{\theta}_{t+1}^{(i)}}, \sigma_{\hat{\theta}_{t+1}^{(i)}}^2, \hat{\theta}_{t+1, \min}^{(i)}, \hat{\theta}_{t+1, \max}^{(i)})$
            \EndFor
            
            \State $\hat{x}_{t+1}, \hat{y}_{t+1}, \hat{z}_{t+1}$ = $FK$(Samples from $\hat{\bm{\theta}}_{t+1}$) \Comment{Use forward kinematics to transform back to task space}
            \State $f_{XYZ} = \prod_{i=1}^{p_n}\hat{\theta}_{i, t+1}\frac{1}{|J|} $ \Comment{Use jacobian method to obtain corresponding pdf in task space}
            \State Reject $\hat{x}_{t+1}, \hat{y}_{t+1}, \hat{z}_{t+1}$ samples that violate constraints and corresponding $f_{XYZ}$.
            \State Normalize constrained $f_{XYZ}$ 
            
        \EndFor

\end{algorithmic}
\end{algorithm}

\section{Implementation and evaluation}
\subsection{Dataset}\label{AA}
To evaluate our approach, we use two relevant datasets that address our requirements: \begin{enumerate}
 \item the ANDY dataset \cite{andydataset} with industry like manual activities,
    \item the HA4M \cite{ha4m} assembly task dataset. 
\end{enumerate} These were selected since it is important to consider datasets in which the human interacts with the surroundings, especially ones that involve the execution of some task or activity. 
The ANDY dataset captures the skeleton data using inertial and optical motion capture sensors recorded at 240Hz and 120Hz respectively. The actions performed are labelled which thus allows the prediction framework to be tested on individual actions and consider workspace constraints like tables, shelves, etc. objects in the scene. On the other hand, the HA4M dataset captures the skeleton data of different subjects performing an assembly task using the Microsoft\textregistered Azure Kinect Camera, where the skeleton joint poses are tracked using the Azure Kinect Body Tracking SDK. The assembly task consists of different actions to build an Epicyclic Gear Train and involves physical constraints to the workspace, example - table where the parts are placed, workspace constraints, etc. 

\subsection{Implementation}

The performance of our framework is demonstrated by using a Gaussian Process Regression (GPR) model for the unconstrained prediction stage as it helps in quantifying the prediction uncertainty. It assumes Gaussian distribution on the noise in the prior i.e. the observed joint angle trajectory. \\

We base our comparison on three methods of varying levels of representation and constraints:

\subsubsection{Unconstrained xyz (GP xyz)}
the hyperparameters of the GPR are tuned on data in which the input representation is past observed xyz joint positions, and the output is the predicted position along with the underlying uncertainty.
\subsubsection{Unconstrained joint angles (GP J.A.)}
the GPR is tuned with inputs in the joint angle space i.e. by applying inverse kinematics (IK) on observed joint positions to obtain joint angles, followed by GPR on the joint angles to obtain predicted joint angles, and propagating these back to xyz space using forward kinematics (FK).

\subsubsection{Constrained prediction (GP Constr.)}
similar to the unconstrained joint angle formulation but now we constrain the joint angle predictions as well as the forward kinematics output xyz values for collision avoidance or intersection with physical objects in the scene.

    For the SPGP, we use a radial basis function (RBF) kernel and use an observed window of 200 milliseconds and a prediction window of 500 milliseconds. We consider trajectories from joints of both hands i.e. shoulder, elbow, and wrist. A simple kinematics model derived from \cite{human_arm_kinematics} and PySwarms solver \cite{pyswarm} for the inverse kinematics is used to get to the joint space, and the joint angle and joint velocity constraints are used as defined in \cite{human_arm_kinematics}.

    \subsection{Simulations}
        
    \textbf{HA4M \cite{ha4m}}: For this particular dataset, we consider left and right arm (shoulder, elbow, and wrist) trajectories from 5 different subjects for training the SPGP and 7 other subjects for evaluating the constraints framework. The frame rate is 30Hz and a step size of 5 frames between every sliding prediction is used and both ends of the trajectories are trimmed to remove extended amounts of stationary behavior. For the constraints, the major one we consider is the table where parts are being picked and placed. In most frames, this was a simple linear constraint in the world z direction of the provided coordinate system. As for velocity constraints, we observe that the wrist motions do not exceed $1.5m/s$ and thus incorporate that as a constraint after transforming it in the joint space. 
    \\
    
    \textbf{ANDY \cite{andydataset}}: This dataset has an additional feature of labelled actions, and thus we evaluate our method only on specific actions - reaching, picking, carrying, and placing. These are the most relevant actions to the application of constrained based prediction. We again use arm trajectories and use a step size of 20 frames instead since the frame rate is 120Hz. 

\subsection{Evaluation Metrics}
We use two metrics to evaluate our work. The first one is the mean per joint position error (MPJPE) on cartesian joint positions
 \begin{equation}\label{eq:mpjpe}
    \mathcal{L}_{\text{MPJPE}} = \frac{1}{N\cdot T_{P}}\sum_{n=1}^{N}\sum_{t = 1}^{T_{P}}\left\lVert  \textbf{j}^{(n)}_{t} - \hat{\textbf{j}}^{(n)}_{t}\right\rVert _2
\end{equation}

as popularly used in other works \cite{sts_gcn}. We use this metric specifically on hand motions i.e. shoulder, elbow and wrist trajectories in order to evaluate interactions with scene constraints. \\ 

The quantified uncertainty is evaluated using the negative log-likelihood (NLL) metric on the ground truth values $\textbf{s}_t$ with respect to the constrained pdf:

\begin{equation}\label{eq:nll}
    \text{NLL} = \frac{1}{T_{P}}\sum_{t = 1}^{T_{P}}-log\big(f_{XYZ}(\textbf{s}_t)\big)
\end{equation}

where the joint pdf $f_{XYZ}$ is derived from Eq. (\ref{eq:pdf}).

\subsection{Results}

As used in previous methods \cite{iplanner}, each GPR model was trained according to a given classified action in the ANDY dataset for each joint trajectory prediction, while on a subset of trajectories of 5 different people from HA4M. Using MPJPE from Eq. (\ref{eq:mpjpe}) on the expectation of the Monte Carlo (MC) based rejection sampled points for each method, we specifically evaluate our method for shoulder, elbow, and wrist trajectories of both arms for selected actions, and results are shown in Table \ref{table:1}.

\begin{table}[htbp]
\centering   

\caption{MPJPE (in mm) using joints - shoulder, elbow, and wrist of both hands. The prediction window is 500ms. Our framework consistently outperforms unconstrained GPs in different representations.}
\label{table:1}
\begin{tabular}{||>{\centering\arraybackslash}p{1cm}|>
{\centering\arraybackslash}p{1cm}|>{\centering\arraybackslash}p{1.3cm}|>{\centering\arraybackslash}p{1.3cm}|>{\centering\arraybackslash}p{1.3cm}||}
 \hline
 Dataset & Action & GP xyz & GP J.A. & GP Constr. \\ [0.5ex] 
 \hline\hline
  \multirow{3}{*}{ANDY} & Reaching & 223 $\pm$ 14 & 87 $\pm$ 5 & \textbf{76 $\pm$ 5} \\ 
 & Picking & 268 $\pm$ 18 & 123  $\pm$ 7 & \textbf{89 $\pm$ 4} \\
 & Placing & 255 $\pm$ 13 & 103 $\pm$ 8 & \textbf{73 $\pm$ 5} \\ \hline
 HA4M & - & 173 $\pm$ 13 & 104 $\pm$ 7 & \textbf{85 $\pm$ 9} \\
 [1ex] 
 \hline
\end{tabular}
\end{table}

It is evident that moving from joint position space to joint angle space followed by adding constraints results in a better MPJPE. We see a considerable reduction in error when changing representation space which can be attributed to both the use of a kinematics model and enforcing constant bone lengths via the same. The next shift is from unconstrained joint space to constrained joint space. This shift is especially noticeable for the actions we tested because of how often we reach the edges of some joint constraints when performing them.

\begin{figure}[htbp]

\centerline{\includegraphics[width=8.4cm, height=4.5cm]{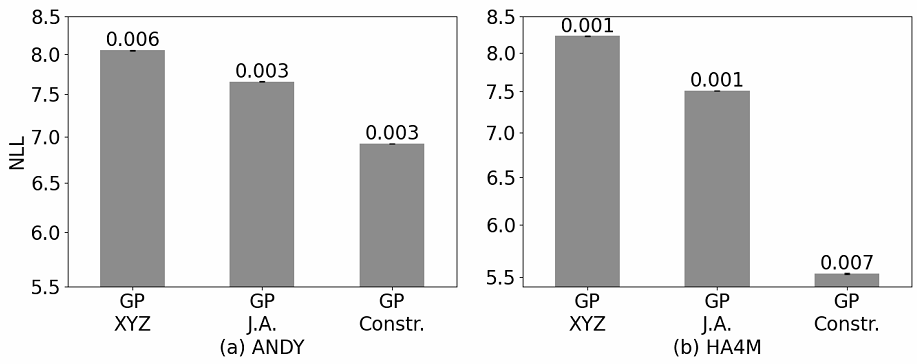}}

\caption{Comparing NLL of ground truth prediction with pdf for ANDY (left) and HA4M (right) for GP xyz, GP joint angles (J.A.) and GP constrained (constr.). Standard errors are on top of the bar plots.}
\label{fig_hist}
\end{figure}

To evaluate the estimated joint pdf, we use the NLL metric from Eq. (\ref{eq:nll}). We run our evaluation for 25 iterations of every method and plot the corresponding histograms on a log scale with the standard error on top of them as shown in Fig. \ref{fig_hist}.

It is again evident that changing representation space and adding constraints perform much better (15\% and 32\% for ANDY and HA4M respectively from Fig. \ref{fig_hist}) as the NLL is lowered, signifying a better probability value for the evaluated ground truth, thus meaning that rejecting points that violated constraints resulted in a considerable improved prediction pdf.

\section{Experiments}

\begin{figure}[htbp]
\centerline{\includegraphics[width=5.5cm, height=4.4cm]{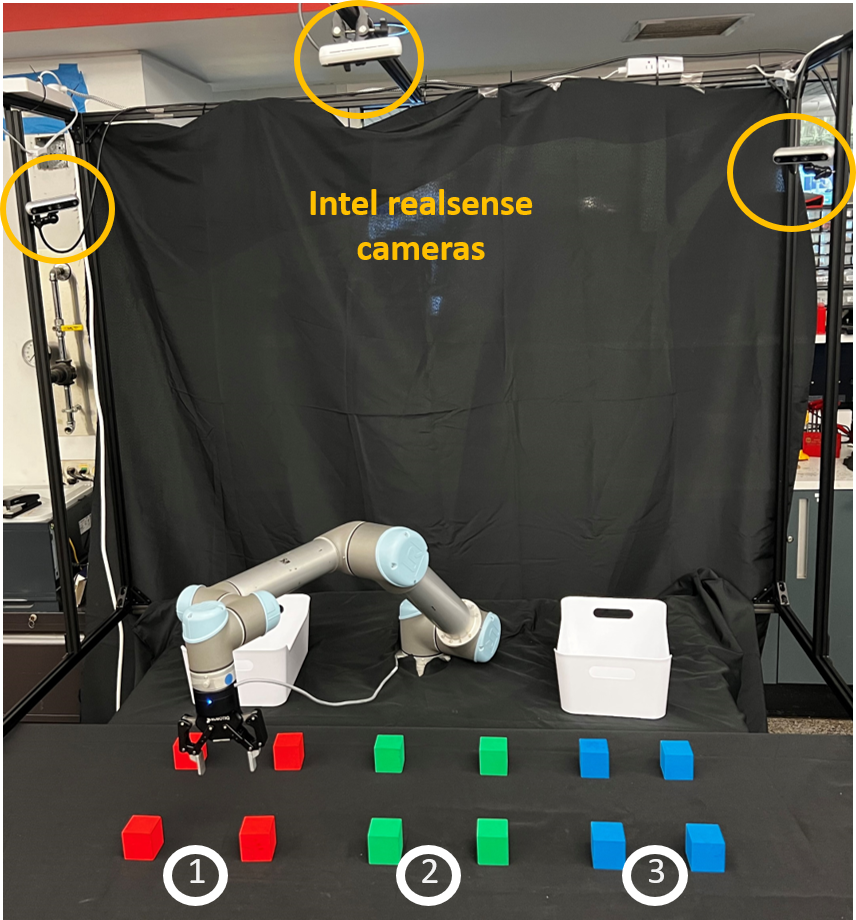}}
\caption{Experimental setup for human-robot collaboration.}
\label{fig:expt_setup}
\end{figure}

\newcommand{\imgwidth}{0.15}
\begin{figure*}[htbp]
\vspace{0.1in}
    \centering
    \begin{minipage}{\imgwidth\textwidth}
        \includegraphics[width=\textwidth]{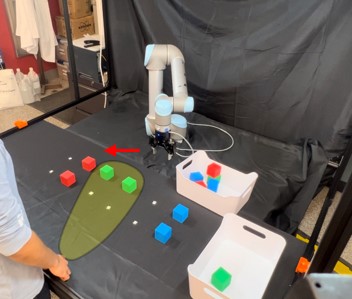}
        \centering{(1a)}
        \label{fig:replan1}
    \end{minipage}
    \hfill
    \begin{minipage}{\imgwidth\textwidth}
        \includegraphics[width=\textwidth]{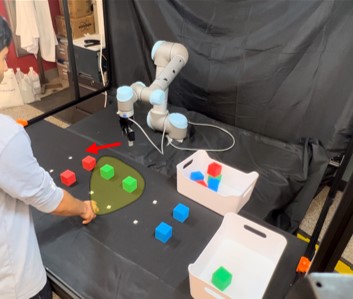}
        \centering{(1b)}
        \label{fig:replan2}
    \end{minipage}
    \hfill
    \begin{minipage}{\imgwidth\textwidth}
        \includegraphics[width=\textwidth]{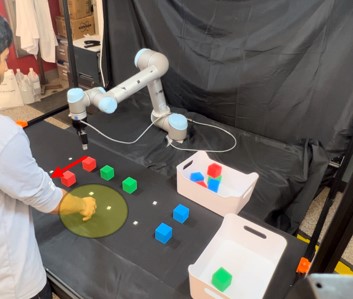}
        \centering{(1c)}
        \label{fig:replan3}
    \end{minipage}
    \hfill
    \begin{minipage}{\imgwidth\textwidth}
        \includegraphics[width=\textwidth]{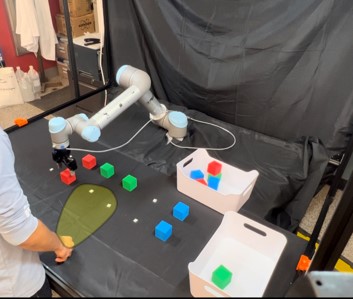}
        \centering{(1d)}
        \label{fig:replan4}
    \end{minipage}
    \hfill
    \begin{minipage}{\imgwidth\textwidth}
        \includegraphics[width=\textwidth]{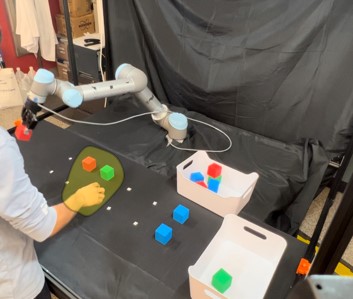}
        \centering{(1e)}
        \label{fig:replan5}
    \end{minipage}

    \begin{minipage}{\imgwidth\textwidth}
        \includegraphics[width=\textwidth]{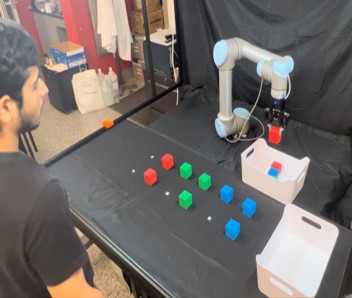}
        \centering{(2a)}
        \label{fig:static_replan1}
    \end{minipage}
    \hfill
    \begin{minipage}{\imgwidth\textwidth}
        \includegraphics[width=\textwidth]{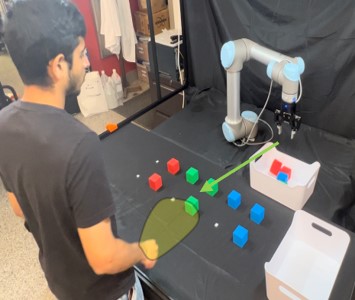}
        \centering{(2b)}
        \label{fig:static_replan2}
    \end{minipage}
    \hfill
    \begin{minipage}{\imgwidth\textwidth}
        \includegraphics[width=\textwidth]{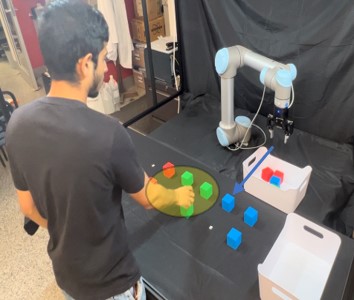}
        \centering{(2c)}
        \label{fig:static_replan3}
    \end{minipage}
    \hfill
    \begin{minipage}{\imgwidth\textwidth}
        \includegraphics[width=\textwidth]{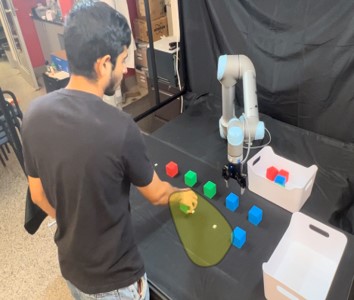}
        \centering{(2d)}
        \label{fig:static_replan4}
    \end{minipage}
    \hfill
    \begin{minipage}{\imgwidth\textwidth}
        \includegraphics[width=\textwidth]{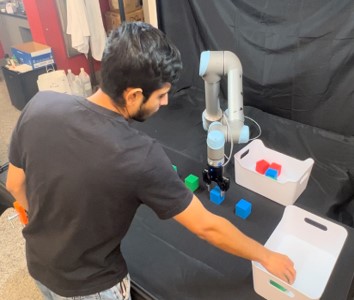}
        \centering{(2e)}
        \label{fig:static_replan5}
    \end{minipage}

    \begin{minipage}{\imgwidth\textwidth}
        \includegraphics[width=\textwidth]{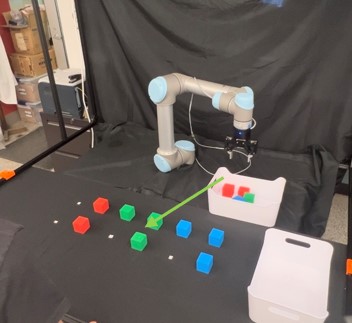}
        \centering{(3a)}
        \label{fig:dynamic_replan1}
    \end{minipage}
    \hfill
    \begin{minipage}{\imgwidth\textwidth}
        \includegraphics[width=\textwidth]{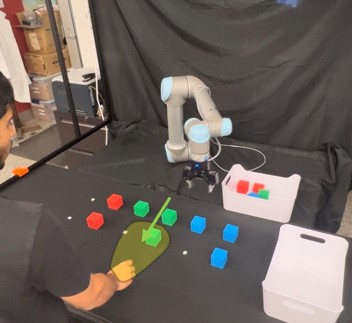}
        \centering{(3b)}
        \label{fig:dynamic_replan2}
    \end{minipage}
    \hfill
    \begin{minipage}{\imgwidth\textwidth}
        \includegraphics[width=\textwidth]{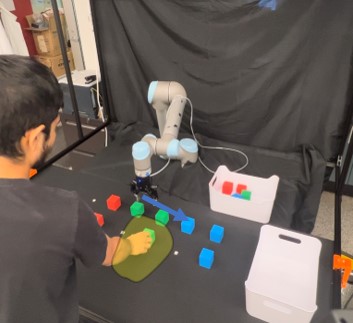}
        \centering{(3c)}
        \label{fig:dynamic_replan3}
    \end{minipage}
    \hfill
    \begin{minipage}{\imgwidth\textwidth}
        \includegraphics[width=\textwidth]{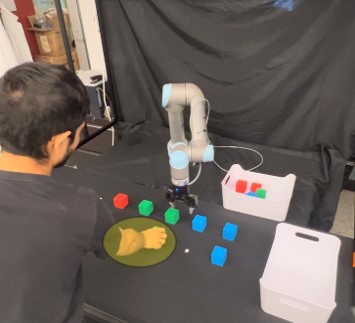}
        \centering{(3d)}
        \label{fig:dynamic_replan4}
    \end{minipage}
    \hfill
    \begin{minipage}{\imgwidth\textwidth}
        \includegraphics[width=\textwidth]{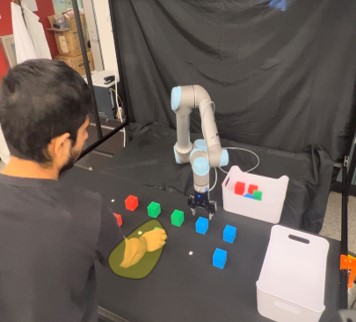}
        \centering{(3e)}
        \label{fig:dynamic_replan5}
    \end{minipage}
    
    \caption{Experimental result. Each row represents a different demonstration where the prediction framework is leveraged for safe and efficient collaboration. The arrows and the corresponding color represent the robot planned motion direction as well as the object to grasp. The yellow region represents the predicted human wrist region. Both arrows and predicted regions are only for qualitative purposes. Row 1 (1a-e) avoiding predicted region - the robot is scheduled to grasp a red cube and the human is reaching out for a green cube; the robot safely navigates around the predicted region while executing the task at hand. Row 2 (2a-e) task re-planning - the robot is scheduled to grasp a green cube but the human predicted region indicates that the human is planning to grasp the same cube; the task is re-planned and the robot plans and grasps a blue block instead. Row 3 (3a-e) task and trajectory re-planning using prediction - similar to 2, the robot is scheduled to grasp a green cube and is executing trajectory for the same. As soon as a human is predicted to be in the green cubes space, the motion is re-planned and the robot grasps a blue cube instead.}
    \label{fig:expt_test}
\end{figure*}

\subsection{Experimental platform}\label{AA}
Fig. \ref{fig:expt_setup} illustrates the experimental setup of a human-robot collaboration task scenario in which a UR5 robot arm is tasked with picking and placing a red, green, and blue cube in order. 

As soon as a human enters the scene, image frames from the Intel RealSense cameras are passed through mmpose \cite{mmpose} to track human joint positions which are fused together and passed to both prediction and planning modules. Our constrained motion prediction framework is used to accurately predict the most probable regions where the human might be in the next 500ms. Depending on the predicted region, the grasp point or order is modified for successful and efficient completion of the task while safely navigating around the human. We use moveit \cite{moveit}, ROS Noetic and CHOMP \cite{chomp} motion planner as the primary motion planner.

\subsection{Demonstration}
To demonstrate the real-time application of the proposed motion prediction framework, three different scenarios are used, as shown in Fig. \ref{fig:expt_test}. The associated video demonstrations can be found at \href{https://www.youtube.com/@MITMechatronics/videos}{youtube.com/@MITMechatronics/videos}.
The overall premise is to pick and place red, green, and blue cubes (in order), one at a time. Using the tracked human joint motions and the prediction framework, certain regions in space become unsafe and the robot has to avoid them, and also in the grasp order, if certain cubes become unsafe, then the grasp order is modified accordingly, under the assumption that the human might manipulate one or more of those cubes. While we only demonstrate one human in the scene, our framework can be successfully applied to multiple humans since for the application, only the predicted occupied region in space is relevant.

\section{Conclusion}

We introduce a novel constrained probability distribution prediction (CPDP) framework for human motion prediction that explicitly accounts for kinematic as well as scene constraints in order to predict more accurate probability distribution functions for a predicted motion trajectory. Our proposed framework is able to reason about the capabilities of the physical system at hand as well as account for any implausible predictions made when predicting the final trajectory pdf. We evaluate our framework on two task-relevant human motion datasets and observe considerable improvements. We also implement it in a real-time human-robot collaboration application using a UR5 robot.

In future work, we intend to make use of CPDP on other existing prediction frameworks like \cite{sts_gcn} to demonstrate the added benefit of our framework to any human motion prediction module for a real-life application.

\section*{Acknowledgment}
This research was made possible by the support and partnership of King Abudlaziz City for Science and Technology (KACST) through the Center for Complex Engineering Systems at Massachusetts Institute of Technology (MIT) and KACST.

\newpage

\bibliographystyle{plain}
\bibliography{main}

\end{document}